\documentclass[letterpaper]{article} 
\usepackage{aaai23}  
\usepackage{times}  
\usepackage{helvet}  
\usepackage{courier}  
\usepackage[hyphens]{url}  
\usepackage{graphicx} 
\usepackage{natbib}  
\usepackage{caption} 
\usepackage{algorithm}
\usepackage{algorithmic}

\usepackage{newfloat}
\usepackage{listings}
\DeclareCaptionStyle{ruled}{labelfont=normalfont,labelsep=colon,strut=off} 
\floatstyle{ruled}
\newfloat{listing}{tb}{lst}{}
\floatname{listing}{Listing}
\usepackage[utf8]{inputenc} 
\usepackage{booktabs}       
\usepackage{amsfonts}       
\usepackage{nicefrac}       
\usepackage{microtype}      
\usepackage[dvipsnames]{xcolor} 

\usepackage{interval}
\usepackage{bm}
\usepackage{bbm}
\usepackage{mathtools}
\usepackage{footmisc}
\usepackage{subfig}
\usepackage{xspace}
\usepackage[inline]{enumitem}

\usepackage{multirow}

\nocopyright

\newcommand*{\eg}{\emph{e.g.}\@\xspace}

\definecolor{xycolor}{RGB}{60, 120, 216}

\definecolor{tcolor}{RGB}{80, 200, 180}

\newlength\savewidth

\definecolor{fastcolor}{RGB}{100,178,100}

\title{Video Mobile-Former: Video Recognition with Efficient \\ Global Spatial-temporal Modeling}
\author {
    Rui Wang\textsuperscript{\rm 1,2},
    Zuxuan Wu\textsuperscript{\rm 1,2},
    Dongdong Chen\textsuperscript{\rm 3},
    Yinpeng Chen\textsuperscript{\rm 3},
    Xiyang Dai\textsuperscript{\rm 3}, \\
    Mengchen Liu\textsuperscript{\rm 3}, 
    Luowei Zhou\textsuperscript{\rm 3},
    Lu Yuan\textsuperscript{\rm 3},
    Yu-Gang Jiang\textsuperscript{\rm 1,2}
}
\affiliations {
    \textsuperscript{\rm 1} Shanghai Key Lab of Intelligent Info. Processing, School of Computer Science, Fudan University, \\
    \textsuperscript{\rm 2} Shanghai Collaborative Innovation Center on Intelligent Visual Computing, 
    \textsuperscript{\rm 3} Microsoft Cloud\&AI\\
    \{ruiwang16,zxwu,ygj\}@fudan.edu.cn, \{cddlyf,zhouluoweiwest\}@gmail.com, \\ \{yiche,xiyang.dai,mengcliu,luyuan\}@microsoft.com
}

\begin{document}

\maketitle

\begin{abstract}
  Transformer-based models have achieved top performance on major video recognition benchmarks. Benefiting from the self-attention mechanism, these models show stronger ability of modeling long-range dependencies compared to CNN-based models. However, significant computation overheads, resulted from the quadratic complexity of self-attention on top of a tremendous number of tokens, limit the use of existing video transformers in applications with limited resources like mobile devices. In this paper, we extend Mobile-Former to Video Mobile-Former, which decouples the video architecture into a lightweight 3D-CNNs for local context modeling and a Transformer modules for global interaction modeling in a parallel fashion. To avoid significant computational cost incurred by computing self-attention between the large number of local patches in videos, we propose to use very few global tokens (e.g., 6) for a whole video in Transformers to exchange information with 3D-CNNs with a cross-attention mechanism. Through efficient global spatial-temporal modeling, Video Mobile-Former significantly improves the video recognition performance of alternative lightweight baselines, and outperforms other efficient CNN-based models at the low FLOP regime from 500M to 6G total FLOPs on various video recognition tasks. It is worth noting that Video Mobile-Former is the first Transformer-based video model which constrains the computational budget within 1G FLOPs.
\end{abstract}

\section{Introduction}

Vision Transformers \cite{vit,liu2021swin,dong2021cswin,dong2021peco} have achieved remarkable progress in computer vision and outperform Convolutional Neural Networks (CNNs) on a multitude of image tasks. By partitioning each image into non-overlapping patches, vision transformers transfer image data to a sequence of patch tokens 
and globally model their relationships with self-attention mechanisms. Following the similar paradigms, recent video transformers \cite{liu2021video,arnab2021vivit,li2022mvitv2,wang2022bevt} extend transformers to the temporal axis and achieve top performance on major video recognition benchmarks. However, video transformers that model spatial-temporal information jointly incur high computational cost, due to the quadratic complexity of self-attention mechanisms for a large number of tokens. Although recent methods reduce the computational burden of video transformers by: 1) decoupling self-attention on the spatial axis and the temporal axis~\cite{timesformer}, or 2) applying window attention~\cite{liu2021video} or pooling attention~\cite{fan2021multiscale,li2022mvitv2} with fewer local tokens, these architectures are still computationally heavy (\eg, more than 100G total FLOPs) and impractical under the resource-constrained scenarios compared to CNN-based models. 

\begin{figure}
    \centering
    \includegraphics[width=0.98\linewidth]{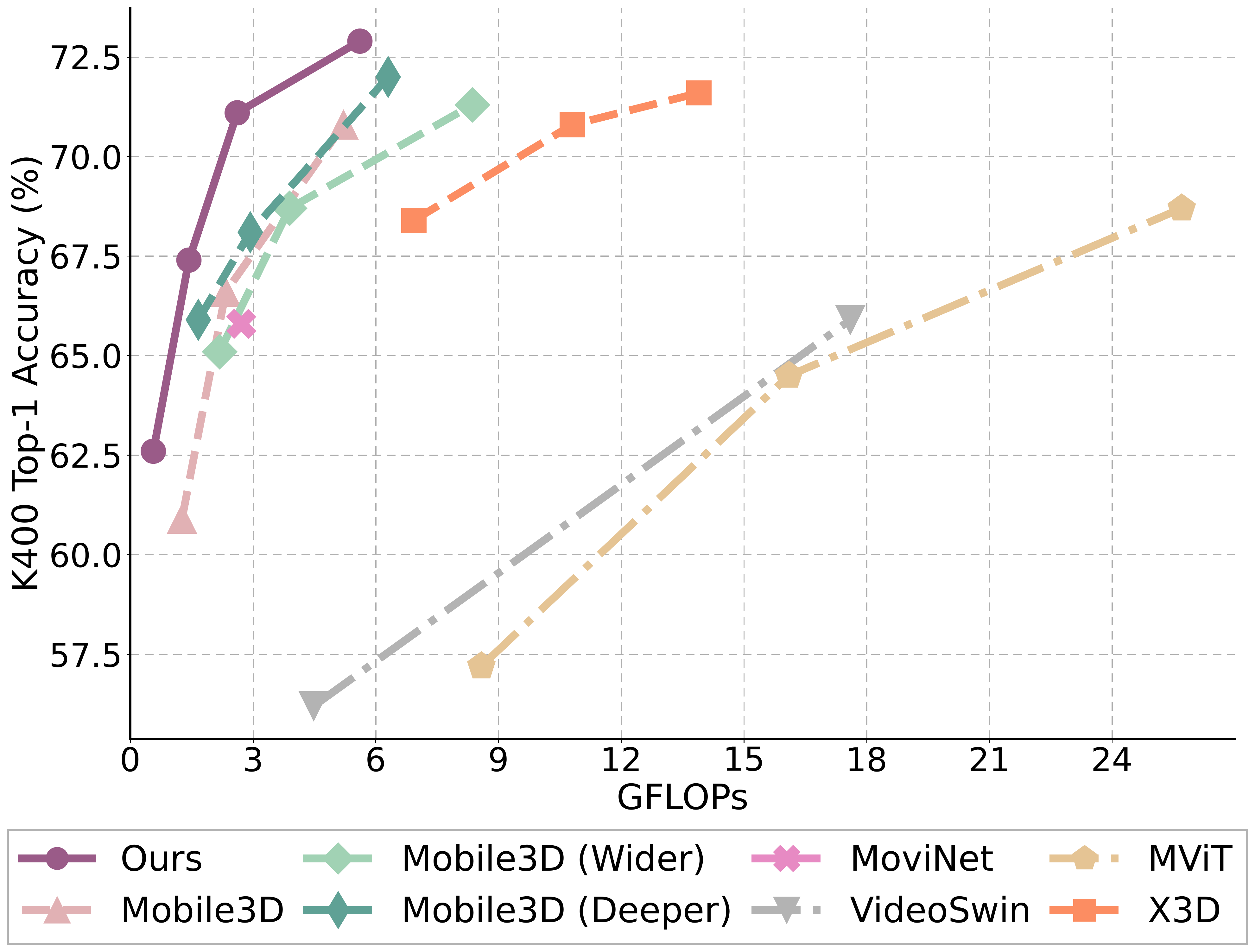}
    \vspace{-0.5em}
    \caption{Comparing Video Mobile-Former with other efficient video models on Kinetics-400. Single-clip evaluation is used here. Video Mobile-Former outperforms both efficient 3D-CNN and video Transformers at the low FLOP regime.}%
    \label{fig:flops_acc_compare}%
    \vspace{-1.5em}
\end{figure}

Among lightweight backbones for video recognition, 3D CNN-based models~\cite{x3d,movinets} that contain an inductive bias of locality are still popular solutions. However, 3D CNNs can not capture global relationships efficiently, as it requires to stack more convolutional layers to achieve the global receptive field. To model global spatial-temporal relationships with yet compact architectures, we aim to design an efficient video transformer at the low FLOPs regime from 500M to 6G total FLOPs. 

\begin{figure}
    \centering
    \includegraphics[width=0.95\linewidth]{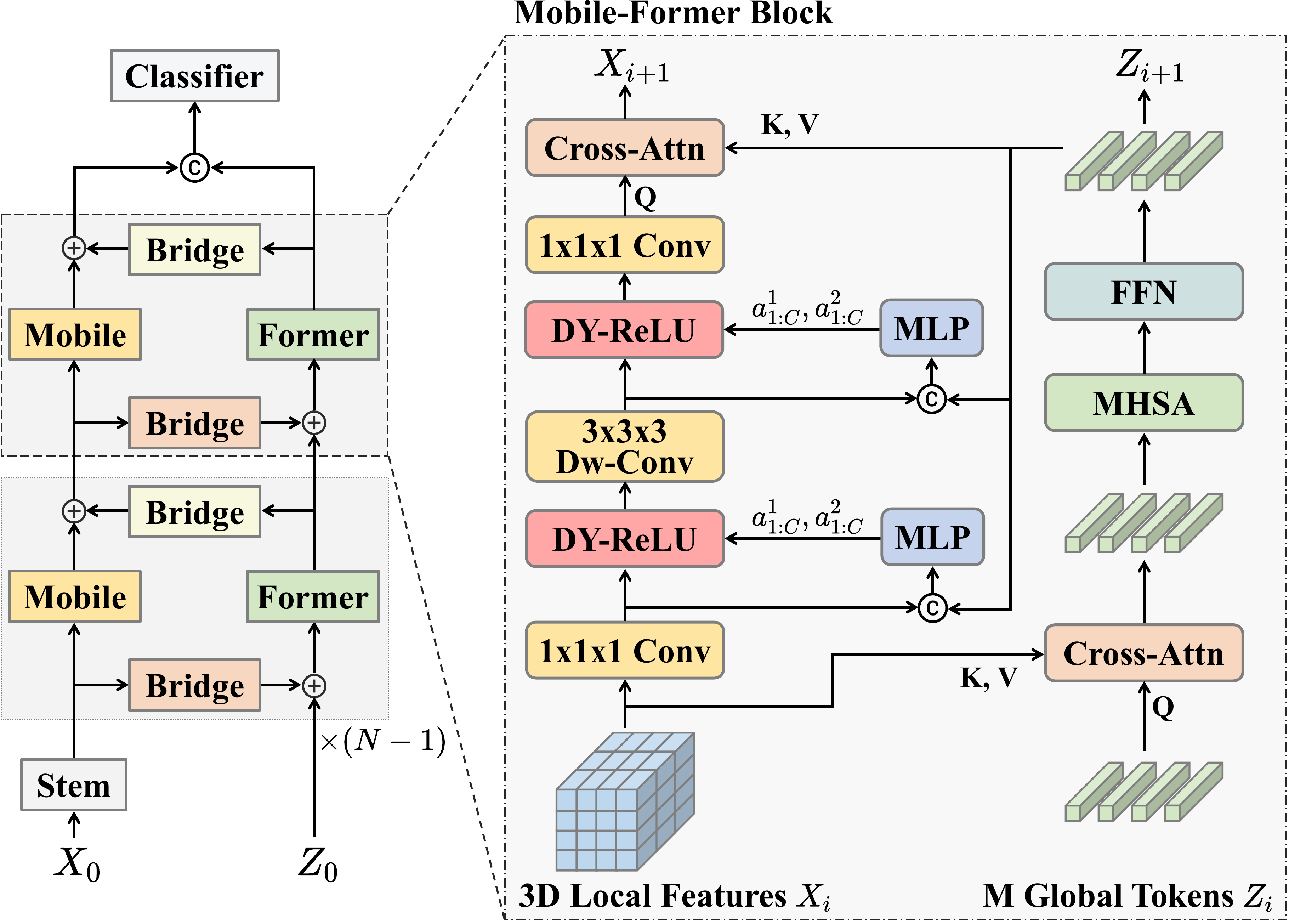}
    \vspace{-0.5em}
    \caption{Left: The overall parallel architecture of Video Mobile-Former. Right: A Video Mobile-Former block including a Mobile block, a Former block and two interaction bridges with cross-attention.}%
    \label{fig:video_mobile_former}%
    \vspace{-1.5em}
\end{figure}

In this paper, we extend Mobile-Former ~\cite{chen2021mobile} from image domain to video domain, dubbed as ``Video Mobile-Former''. As shown in Figure \ref{fig:video_mobile_former}, it decouples the video architecture into a lightweight 3D-CNN for local context modeling and a Transformer module for global interaction modeling. A Video Mobile-Former block is a two-pronged parallel structure consisting of a Mobile branch, a Former branch, and two interaction bridges. The Mobile branch is based on the lightweight 3D-CNN block of the ``mobile-regime'' models~\cite{x3d}, which is composed of the 3D depthwise and pointwise convolution layers. The Former branch, on the other hand, consisting of multi-head self-attention modules (MHSA) and feed-forward networks (FFNs), aims to capture global relationships by applying self-attention on several learnable global tokens. To exchange information between Mobile branches and Former branches, two interaction bridges are introduced by applying cross-attention between global tokens and local features of 3D-CNN blocks. Bidirectional communication between local context modeling modules with the \textit{Mobile branch} and global relationship modeling modules with the \textit{Former branch} enhances the output features of Video Mobile-Former blocks, which leads to stronger spatial-temporal representations and better video recognition performance.

Our main contribution is an efficient Transformer-based video model that has an extremely low computational cost and a stronger ability to extract spatial-temporal features compared to lightweight CNN-based models.
While generalizing a top-performing image recognition network to the video domain is appealing, it is non-trivial as the additional time dimension contains significant redundancy and poses a great challenge for temporal modeling. It is often found that directly extending 2D networks to 3D produces limited results~\cite{karpathy2014large}. This motivates us to study how to better adapt image architectures to video architectures. In particular, we leverage three important features in our architectural design: (a) In Video Mobile-Former, global features of a whole video can be extracted with very few global tokens (\eg, less than six tokens). Using more global tokens or applying $M$ global tokens for each frame (which means applying self-attention on $M*T$ tokens for a video of $T$ frames) fails to produce significant gains. Former branches and interaction bridges consume less than 12\% of total computation since they use extremely few tokens compared with previous video transformers. (b) In the convolutional stem at the bottom of our network, a good trade-off between performance and computational cost can be achieved using a convolutional layer with a large temporal stride, which temporally downsamples the input videos. (c) Due to the global spatial-temporal modeling performed by Former modules, Video Mobile-Former can outperform deeper 3D-CNNs with a shallow network, which significantly reduces the computational cost.

We evaluate our Video Mobile-Former on several video recognition benchmarks. Experiment results demonstrate that Video Mobile-Former outperforms state-of-the-art lightweight CNN-based models and Video Transformers at the low FLOP regime from 500M to 6G total FLOPs. In comparison to alternative lightweight baselines that only utilize Mobile branches, Video Mobile-Former significantly improves the video recognition performance. Additionally, enhancing 3D-CNNs with Former branches of Video Mobile-Former is more effective and more efficient compared to widening or deepening the networks. 

\section{Related Work}
\vspace{0.05in}
\noindent \textbf{Video Recognition with Efficient CNNs. } 
To explore the temporal information of video data, a popular strategy for designing video recognition models is to extend 2D image networks on the temporal dimension. Therefore, deep 3D CNNs adapted from 2D CNNs are widely used~\cite{c3d,i3d,p3d,r21d,tpn,eco,channelseparated,slowfast}. Due to the increasing computational cost and parameters of 3D CNNs, improving the efficiency of video models has become the spotlight of video recognition. Several efficient architectures of 3D CNNs~\cite{kopuklu2019resource,x3d,movinets} leverage the idea of efficient 2D CNNs~\cite{iandola2016squeezenet,mobilenets,mobilenetv2,zhang2018shufflenet,ma2018shufflenetv2}. Efficient video architectures can also be derived from Neural Architecture Search (NAS)~\cite{movinets,tinyvideonet} . Another trend of developing efficient video models is to integrate lightweight temporal modules with 2D CNNs~\cite{tsn,tsm,tam,videolstm,kwon2020motionsqueeze}. Despite the efficiency, such CNN-based solutions still struggle with modeling the global dependency, which is required by some video recognition tasks that need global understanding. 

\vspace{0.05in}
\noindent \textbf{Efficient Vision Transformers. }
Transformers~\cite{vaswani2017attention} are firstly proposed for natural language processing and is introduced into vision tasks recently. Vision Transformer (ViT)~\cite{vit} and its subsequent variants~\cite{liu2021swin,dong2021cswin,wang2021pyramidvit} achieve advanced performance on various vision tasks by customizing the designs for the vision signal. However, compared to CNNs, ViTs have high computational complexity and a large number of parameters, which limits their application in resource-constrained scenarios. To address this issue, researchers combine the advantages of ViTs and CNNs~\cite{chen2021mobile,li2022mvitv2,mobilevit,zhou2021elsa,xu2021vitae}, or develop efficient self-attention mechanisms for separated local-global relationship modeling~\cite{chu2021twins,el-nouby2021xcit}. To further improve the efficiency of ViTs, recent studies focus on  constructing lightweight transformer-based model architectures for on-device applications~\cite{chen2021mobile,mobilevit,mobilevitv2,pan2022edgevits,li2022efficientformer,zhang2022edgeformer}. Different from these efficient image transformers, we aim to design an efficient video transformer by considering the high redundancy existing in videos.

\vspace{0.05in}
\noindent \textbf{Video Transformers. } 
The success of ViT has ignited growing interest in leveraging the ability of transformers to model the long-term dependencies for video recognition tasks. To reduce the computational cost, variants of self-attention mechanisms are introduced~\cite{arnab2021vivit,timesformer,liu2021video,bulat2021space,fan2021multiscale,zha2021shifted,motionformer,multiview_transformer}. TimeSformer~\cite{timesformer} divides the spatial-temporal modeling into temporal attention and spatial attention that are separately computed in each layer. ViViT~\cite{arnab2021vivit} studies several variants of space-time factorization for video transformers. VideoSwin~\cite{liu2021video} integrates an inductive bias of locality with transformer, where self-attention is computed in a 3D local window. MViT~\cite{fan2021multiscale,li2022mvitv2}, a hierarchical architecture, applies pooling self-attention that significantly reduces the token number. To approximate the space-time attention with relatively low computational complexity, X-ViT~\cite{bulat2021space} restricts time attention to a local temporal window and employs an efficient space-time mixing mechanism. TokenLearner~\cite{tokenlearner} uses a plug-in module that extracts a few important visual tokens for each video frame and learns video representation with pairwise attention between such tokens. Uniformer~\cite{li2022uniformer} utilizes 3D CNN and spatial-temporal self-attention respectively in shallow and deep layers. Compared with CNN-based models, transformer-based models achieve remarkable progress on video recognition tasks. However, existing video transformers are still computationally heavy, and lightweight transformers for video recognition have not been studied yet. 

\section{Methodology}

\subsection{Revisiting Mobile-Former}

In this paper, we introduce a lightweight video transformer by extending Mobile-Former architecture that is designed for image tasks. Mobile-Former is a parallel architecture bridging MobileNet~\cite{mobilenets,mobilenetv2} and Transformer~\cite{vit} with bidirectional cross-attention. Mobile-Former consists of stacked Mobile-Former blocks. Each block includes a Mobile block, a Former block and two interaction bridges. 

A Mobile block uses an inverted bottleneck block of MobileNet V2~\cite{mobilenetv2} to extract local features. The Mobile block of the $i^{th}$ layer takes the feature map $X_i \in \mathbb{R}^{H_i \times W_i \times C_i} $ as inputs, which is processed by the $1\times1$ pointwise convolution and the $3\times3$ depthwise convolution later. 
A Former block is a transformer module aiming to compute global relations of input features, which includes a multi-head attention (MHA) and a feed-forward (FFN). To reduce the computational cost, Former block takes $M$ learnable global tokens $\bm{Z} \in \mathbb{R}^{M \times d}$ instead of dense patch tokens as input. The input of attention is split as $\bm{Z} = [\bm{Z}_s]$ for MHA with $H$ heads. The simplified self-attention of multiple global tokens is defined as:
\begin{align}
\mathcal{A}_{\bm{Z}}&=\left[Attn(\bm{Z}_s\bm{W}_s^Q, \bm{Z}_s, \bm{Z}_s)\right]_{s=1:H}\bm{W}^O,
\label{eq:former-block}
\end{align}
where $\bm{W}_s^Q$ is the projection matrix of query in the $s^{th}$ head, $\bm{W}^O$ aims to combine the channel information of different heads. And $Attn$ denotes the standard attention:
 \begin{align}
Attn(Q, K, V) = softmax(\frac{QK^T}{\sqrt{D}})V.
\label{eq:attn}
\end{align}
 The computational complexity of the Former block is $O(M^2d + Md^2)$.

To exchange information between the Mobile block and the Former block, cross-attention is applied between local features and global tokens. We denote the reshaped local feature map as $\bm{\tilde{X}} \in \mathbb{R}^{\tilde{L}_i \times C_i}$ where $\tilde{L}_i = H_i \times W_i$. The cross-attention from local features to global tokens follows:
 \begin{align}
\mathcal{A}_{\bm{X}\rightarrow\bm{Z}}&=\left[Attn(\bm{Z}_s\bm{W}_s^Q, \bm{\tilde{X}}_s, \bm{\tilde{X}}_s)\right]_{s=1:H}\bm{W}^O.
\label{eq:mobile2former}
\end{align}
Then the derived global tokens are sent to the Former block. After computing the global relations, the cross-attention from global tokens to local features is defined as:

\begin{align}
\mathcal{A}_{\bm{Z}\rightarrow\bm{X}}&=\left[ Attn(\bm{\tilde{X}}_s, \bm{Z}_s\bm{W}_s^K, \bm{Z}_s\bm{W}_s^V) \right]_{s=1:H},
\label{eq:Former2Mobile}
\end{align}
where global information is injected to the output of the Mobile block, and $\bm{W}_s^K$ and $\bm{W}_s^V$ are the projection matrices for the key and the value. In these two cross-attention modules, two policies are applied to
reduce the computational cost: 1) taking local features at the bottleneck of Mobile block as the input of cross-attention due to fewer channels; b) removing the projection matrices from the side of the local features where the length of the input sequence is long, but keeping them at the side of global tokens. 

In Mobile-Former, since the number of global tokens $M$ is no more than six, the Former block and cross-attention modules are computational efficient while the Mobile block consumes the most computation. In our work, we focus on adapting Mobile-Former from the image domain to the video domain and below we discuss how to extend the Mobile CNN branch from 2D to 3D and how to modify the Former branch for efficient temporal modeling. While these extensions sound straightforward, they are non-trivial as the additional temporal axis brings not only more information to process but also more redundancy as well.

\begin{figure*}%
    \centering
    \subfloat[\centering Naive temporal extension of transformers from 2D to 3D. ]{\includegraphics[width=0.98\linewidth]{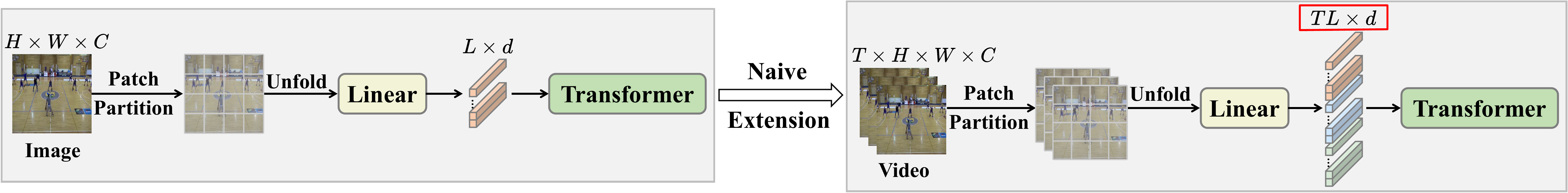} }%
    \qquad
    \subfloat[\centering  Efficient extension of transformer-based modules from Mobile-Former to Video Mobile-Former.]{\includegraphics[width=0.98\linewidth]{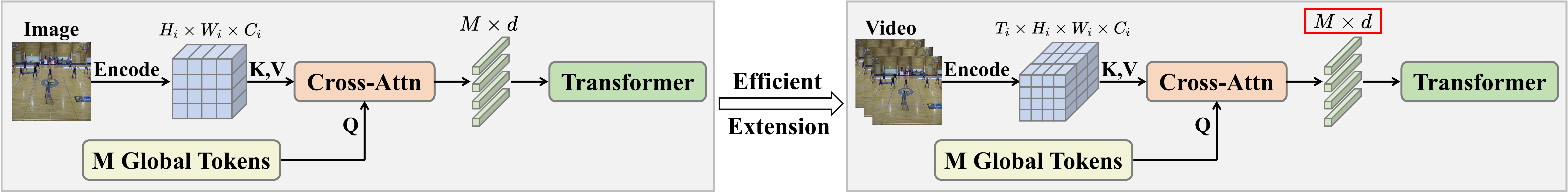} }%
    \vspace{-0.5em}
    \caption{Temporal extension of transformers in previous work and efficient extension of transformer-based modules in our work. }%
    \label{fig:extension_compare}%
    \vspace{-1.5em}
\end{figure*}

\subsection{Extension of CNN from 2D to 3D}

To capture temporal information, 3D convolution is widely used in previous literature of video recognition. One effective method of constructing CNN-based video models is to extend existing image architectures by replacing 2D convolutions with 3D convolutions. In 2D CNN models, the spatial dimension of the feature map is substantially downsampled. However, in most recent 3D CNN models, the temporal dimension of video features is not downsampled~\cite{x3d,movinets}, or is progressively downsampled~\cite{c3d,i3d}. When the number of input video frames is large, the computational cost of the video model to encode a video clip will be much larger than that of the image model to encode an image.  

In Video Mobile-Former, to make an efficient adaption of Mobile block, we introduce large temporal downsampling that reduces the size of feature maps at the early stage of the video model. Specifically, for the input video clip $X \in \mathbb{R}^{T \times H \times W \times 3} $, we use large temporal stride $s_t$ ($s_t \geq 4$) in the 3D convolution to perform temporal downsampling at the stem of Video Mobile-Former, which reduces the temporal length of the output feature map to $T / s_t$. Therefore, the computational cost of the following Mobile blocks and the cross-attention blocks is reduced by a factor of $s_t$. For Mobile block, we replace 2D $3 \times 3$ depthwise convolutions and 2D $1 \times 1$ pointwise convolutions with 3D $3 \times 3 \times 3$ depthwise convolutions and 3D $1 \times 1 \times 1$ pointwise convolutions respectively. The computational complexity of Mobile block is $O(THWC^2/s_t)$.

\subsection{Extension of Transformer from 2D to 3D}

Most video Transformers are extended from the image Transformers in the temporal dimension, and perform global spatial-temporal modeling for the input video 
~\cite{fan2021multiscale,arnab2021vivit}. The longer the input token sequence length is, the heavier the computational cost. For image Transformers, the token sequence with a length of $L_i = H_i \times W_i$, generated by reshaping the image feature map, is processed by MHSA. When extended to the video domain, the input of the Transformer blocks is converted from 2D image features to 3D video features, and MHSA needs to process the tokens sequence of length $T_i \times H_i \times W_i$ after the video features are reshaped. Since the computational complexity of SA grows quadratically with respect to the sequence length of the input, the extension of MHSA in the temporal dimension will incur a very large increase in computation. 

Although some video transformers perform self-attention through pooling ~\cite{li2022mvitv2} or within a local window~\cite{liu2021video}, for video data, the length of the input token sequence is very long because it is proportional to the number of video frames, thus the huge computational cost of transformers is still a significant problem and cannot meet the extremely low FLOPs requirement for devices.

To extend Former block and the interaction bridge from 2D to 3D, if we follow the previous paradigm of extending image models and use $M$ global tokens for each frame ($X_{i,t} \in \mathbb{R}^{H_i \times W_i \times C_i}$ at the $i^{th}$ layer) in the interaction bridge, we would obtain $T \times M$ global tokens for the input of Former block. This strategy is also used in TokenLearner~\cite{tokenlearner}. However, we find this will not only incur non-negligible cost compared to the Mobile block but also is unnecessary.

\begin{table}
\begin{center}
\small
\resizebox{\linewidth}{!}{
\setlength{\tabcolsep}{0.3mm}{

\begin{tabular}{l@{\hspace{3pt}}|c@{\hspace{3pt}}|c|c|c|c|c}
    \toprule
        Models & \#Blocks & \#Dim & \#Exp & \#Res & \#Param & \#FLOPs \\
    \midrule
        Ours-560M & 1,2,3,2 & 16 & 2.25 & 172 & 4.7M & 0.56G \\
        Ours-1G & 1,2,3,2 & 16 & 2.25 & 224 & 4.7M & 1.43G \\
        Ours-2G & 1,2,4,2 & 20 & 2.25 & 224 & 6.9M & 2.61G \\
        Ours-5G & 1,2,12,4 & 24 & 3 & 224 & 13.8M & 5.61G \\
    \bottomrule
    \end{tabular}}}
    \vspace{-0.5em}
    \caption{Detailed configurations of different variants of Video Mobile-Former. The FLOPs are calculated with 64 frames.}
  \label{tab:variants_arch}
  \vspace{-2.5em}
 \end{center}
\end{table}

On account of the frame redundancy in videos, in the Video Mobile-Former, we extract the global information of one whole video with a small constant number of tokens $\bm{Z} \in \mathbb{R}^{M \times d}$ ($M \leq 6$ in this paper), which is not related to the video length. Therefore, in the interaction bridge between Mobile block and Former block, all the global tokens communicate with the feature map of all the frames by the cross-attention. The formulation of the cross-attention in Video Mobile-Former is similar to that in the Mobile-Former but the token length of the reshaped feature map $\bm{\tilde{X}}$ is $T/s_t \times H \times W$. And the computational complexity of interaction bridges is $O(THWMC/s_t + MdC)$. 

In the Former block, MHSA is performed on only $M$ global tokens to obtain global relationships of the whole video. The computational complexity of Former block is $O(M^2d + Md^2)$, which means the computational cost of Former block for a video is equal to that for an image.

In Mobile-Former, ReLU in the Mobile block is replaced by dynamic ReLU~\cite{chen2020dynamicrelu}, the parameters of which are generated by applying two MLP layers on the first global token output from the Former block. In the Video Mobile-Former, to introduce the temporal difference of each frame, we propose \textit{frame-level dynamic ReLU}. Specifically, we concatenate the average-pooled feature of each frame (the frame-level feature) with the first global tokens, and the concatenated feature (the shape is $(T/s_t \times 1 \times 1 \times (C+d))$) is taken as the input of two MLP layers to generate the parameters of dynamic ReLU. Therefore, the activation functions on different temporal positions are different in the Mobile block of Video Mobile-Former.

\subsection{Network Specification}

As shown in Table~\ref{tab:variants_arch}, we build four Video Mobile-Former variants with different computational costs from 560M to 5.61G FLOPs, and refer them by their FLOPs, e.g. Video-Mobile-Former-560M, Video-Mobile-Former-1G. We design different variants by changing the depth of the model (\#Blocks), the base channel number (\#Dim) and the expansion ratio (\#Exp) of each Mobile block, while all blocks have six global tokens with the same dimension 128. All variants have four stages and we employ spatial downsampling (stride $=2$) at the beginning of each stage. Note that we set temporal stride $s_t = 8$ at the stem of Video-Mobile-Former-560M and set $s_t = 4$ at the stem of other variants. The spatial resolution of input video (\#Res) is 172 for Video-Mobile-Former-560M and 224 for other variants.

\section{Experiments}

\noindent \textbf{Datasets.} 
We conduct extensive experiments on Kinetics-400~\cite{kay2017kinetics} and Kinetics-600~\cite{carreira2018short}. Kinetics datasets contain video clips from YouTube with an average duration of 10 seconds. While for Kinetics-400 with 400 categories, we use ~246K videos for training and ~20K videos for testing; for Kinetics-600 with 600 categories we use ~392K videos for training and ~20K videos for testing. We also compare the smallest variant of Video Mobile-Former with efficient 3D-CNN models on Diving48~\cite{diving48} with ~17K fine-grained diving sequences, UCF-101~\cite{soomro2012ucf101} with 13320 videos from 101 action categories, and HMDB-51~\cite{kuehne2011hmdb} with 6849 videos from 51 action categories. 

\vspace{0.05in}
\noindent \textbf{Implementation Details.} 
All models are trained from scratch for Kinetics datasets, while for Diving-48, UCF-101 and HMDB-51, we utilize the model weights pretrained on the Kinetics-600. On account of the balance of training time and performance, for Video Mobile-Former from 1G to 5G, we train with 32 frames, while we train Video Mobile-Former-560M with 64 frames due to the larger temporal stride. For inference, we evaluate all models with a single clip of 64 frames that covers the whole video. As mentioned in \cite{movinets}, single-clip evaluation achieves the balance between the computational cost and performance, compared with multi-clip evaluation. To compare with previous video models fairly, we report the total FLOPs per video, which is a hardware-independent metric.


\begin{table}[t]
  \centering
\setlength{\tabcolsep}{0pt} 
  \begin{tabular*}{\linewidth}{@{\extracolsep{\fill}}lccccc@{}}
    \toprule
    Model & Top-1 & Top-5 & GFLOPs & Frames \\ 
    \midrule
    Mobile3D-1G & 60.9 & 83.7 & 1.26 & 1 $\times$ 64 \\ 
    Mobile3D-1G (wider) & 65.1 & 86.7 & 2.18 & 1 $\times$ 64 \\ 
    Mobile3D-1G (deeper) & 65.9 & 86.8 & 1.66 & 1 $\times$ 64 \\ 
    Video-Mobile-Former-1G & \textbf{67.4} & 87.4 & 1.43 & 1 $\times$ 64 \\ 
    \midrule
    Mobile3D-2G & 66.6 & 87.2 & 2.33 & 1 $\times$ 64 \\ 
    Mobile3D-2G (wider) & 68.7 & 88.6 & 3.89 & 1 $\times$ 64 \\ 
    Mobile3D-2G (deeper) & 68.1 & 88.4 & 2.93 & 1 $\times$ 64 \\ 
    Video-Mobile-Former-2G & \textbf{71.1} & 89.3 & 2.61 & 1 $\times$ 64 \\ 
    \midrule
    Mobile3D-5G & 70.8 & 89.8 & 5.21 & 1 $\times$ 64 \\ 
    Mobile3D-5G (wider) & 71.3 & 89.9 & 8.36 & 1 $\times$ 64 \\ 
    Mobile3D-5G (deeper) & 72.0 & 90.3 & 6.30 & 1 $\times$ 64 \\ 
    Video-Mobile-Former-5G  & \textbf{72.9} & 90.0 & 5.61 & 1 $\times$ 64 \\ 
    \bottomrule
  \end{tabular*}
  \vspace{-0.5em}
    \caption{Comparison to CNN baselines (Mobile3D) and the extension of CNN baselines on Kinetics-400, where ``wide'' and ``deeper'' means widened/deepened Mobile3D baselines.}
  \label{tab:compare_cnn}
  \vspace{-1em}
\end{table}

\begin{table}[t]
  \centering
\setlength{\tabcolsep}{0pt} 
  \begin{tabular*}{\linewidth}{@{\extracolsep{\fill}}lccccc@{}}
    \toprule
    Model & Top-1 & Top-5 & GFLOPs & Frames \\ 
    \midrule
    VideoSwin-14G & 42.4 & 69.9 & 13.9 & 12 $\times$ 32 \\ 
    MViT-27G & 56.0 & 80.7 & 26.8 & 5 $\times$ 16 \\ 
    Video-Mobile-Former-560M & \textbf{62.6} & 83.7 & \textbf{0.56} & 1 $\times$ 64 \\ 
    \midrule
    MViT-43G & 62.0 & 84.8 & 42.9 & 5 $\times$ 16 \\ 
    VideoSwin-54G & 62.2 & 85.0 & 53.8 & 12 $\times$ 32 \\ 
    Video-Mobile-Former-1G & \textbf{67.4} & 87.4 & \textbf{1.43} & 1 $\times$ 64 \\ 
    \midrule
    MoviNet-A0 & 65.8 & 87.4 & 2.71 & 1 $\times$ 50 \\ 
    Video-Mobile-Former-2G & \textbf{71.1} & 89.3 & \textbf{2.61} & 1 $\times$ 64 \\ 
    \midrule
    MobileNetV2+TSM & 69.5 & 88.7 & 72.0 & 30 $\times$ 8 \\ 
    X3D-XS & 69.5 & - & 23.3 & 30 $\times$ 4 \\ 
    MViT-80G & 69.8 & 89.2 & 80.3 & 5 $\times$ 16 \\ 
    ARTNet & 70.7 & 89.3 & 5875 & 250 $\times$ 16 \\ 
    X3D-S & 70.8 & 89.8 & 9.75 & 1 $\times$ 50 \\
    I3D & 71.1 & 90.3 & 108 & 1 $\times$ 250 \\ 
    VideoSwin-212G & 71.7 & 90.3 & 212 & 12 $\times$ 32 \\ 
    R(2+1)D-RGB & 72.0 & 90.0 & 1520 & 10 $\times$ 32 \\ 
    MoviNet-A1 & 72.7 & 91.2 & 6.02 & 1 $\times$ 50 \\ 
    Video-Mobile-Former-5G & \textbf{72.9} & 90.0 & \textbf{5.61} & 1 $\times$ 64 \\ 
    \bottomrule
  \end{tabular*}
    \vspace{-0.5em}
    \caption{Comparison to state-of-the-art on Kinetics-400.}
  \label{tab:k400}
  \vspace{-1.5em}
\end{table}

\begin{table*}[t]
  \centering
\setlength{\tabcolsep}{0pt} 
  \begin{tabular*}{\linewidth}{@{\extracolsep{\fill}}lcccccc@{}}
    \toprule
    Model & MFLOPs & Param & K600 & Diving48 & UCF-101 & HMDB51 \\
    \midrule
    3D-ShuffleNetV1 2.0x & 393  & 4.78M & 56.8 & 41.1 & 85.0 & 53.3 \\
    3D-ShuffleNetV2 2.0x & 360  & 6.64M & 55.2 & 39.4 & 83.3 & 53.5 \\
    3D-MobileNetV1 2.0x & 454  & 14.1M & 48.5 & 34.6 & 76.2 & 46.5 \\
    3D-MobileNetV2 1.0x & 446  & 3.12M & 50.7 & 45.3 & 81.6 & 48.2 \\
    3D-SqueezeNet & 728  & 2.15M & 40.5 & 38.9 & 74.9 & 43.1 \\
    Video-Mobile-Former-560M & 560 & 4.7M & \textbf{65.8} & \textbf{67.8} & \textbf{92.9} & \textbf{70.8} \\
    \bottomrule
  \end{tabular*}
    \vspace{-0.5em}
    \caption{Comparison to resource-efficient 3D-CNN models~\cite{kopuklu2019resource} on Kinetics-600 and other downstream tasks. }
  \label{tab:downstream}
  \vspace{-1.5em}
\end{table*}

\begin{table}[t]
  \centering
\setlength{\tabcolsep}{0pt} 
  \begin{tabular*}{\linewidth}{@{\extracolsep{\fill}}lccccc@{}}
    \toprule
    Model & Top-1 & Top-5 & GFLOPs & Frames \\ 
    \midrule
    MViT-27G & 63.7 & 86.3 & 26.8 & 5 $\times$ 16 \\ 
    Video-Mobile-Former-560M & \textbf{65.8} & 86.3 & \textbf{0.56} & 1 $\times$ 64 \\ 
    \midrule
    MViT-43G & 65.1 & 87.1 & 42.9 & 5 $\times$ 16 \\ 
    Video-Mobile-Former-1G & \textbf{71.1} & 90.0 & \textbf{1.43} & 1 $\times$ 64 \\ 
    \midrule
    X3D-XS & 70.2 & - & 3.88 & 1 $\times$ 20 \\ 
    MoviNet-A0 & 71.5 & 90.4 & 2.71 & 1 $\times$ 50 \\ 
    Video-Mobile-Former-2G & \textbf{74.3} & 91.4 & \textbf{2.61} & 1 $\times$ 64 \\ 
    \midrule
    I3D & 71.9 & 90.1 & 108 & 1 $\times$ 250 \\ 
    X3D-XS & 72.3 & - & 23.3 & 30 $\times$ 4 \\ 
    MViT-80G & 73.2 & 91.6 & 80.3 & 5 $\times$ 16 \\ 
    X3D-S & 74.3 & - & 9.75 & 1 $\times$ 50 \\ 
    MoviNet-A1 & 76.0 & 92.6 & 6.02 & 1 $\times$ 50 \\ 
    Video-Mobile-Former-5G & \textbf{76.4} & 92.7 & \textbf{5.61} & 1 $\times$ 64 \\ 
    \bottomrule
  \end{tabular*}
    \vspace{-0.5em}
    \caption{Comparison to state-of-the-art on Kinetics-600.}
  \label{tab:k600}
  \vspace{-1.5em}
\end{table}

\subsection{Main Results}

\noindent \textbf{Effectiveness and efficiency of Video Mobile-Former.}  
Video Mobile-Former is a parallel architecture including CNN modules and transformer-based modules, which can also be seen as strengthening the CNN-based model with global spatial-temporal modeling. Standard methods of strengthening CNN-based models include widening or deepening the networks. For all variants of the Video Mobile-Former, we obtain the corresponding Mobile3D baselines by removing the Former blocks and the interaction bridges from them, then we compare Video Mobile-Former with the widened baseline and the deepened baseline that have similar/slightly larger FLOPs. As shown in Table~\ref{tab:compare_cnn}, Video Mobile-Former achieves significant higher Top-1 accuracy compared with Mobile3D baselines, while consuming slightly more FLOPs. Additionally, Video Mobile-Former, with less FLOPs, outperforms widened or deepened Mobile3D baselines. This demonstrate the effectiveness and the efficiency of Video Mobile-Former.

\vspace{0.05in}
\noindent \textbf{Comparisons with state-of-the-art efficient video models.}  
In Table~\ref{tab:k400} and ~\ref{tab:k600}, we compare Video Mobile-Former with state-of-the-art efficient video models, including CNN-based models and Transformer-based models, at the low FLOP regime from 500M to 6G total FLOPs. Compared with MoviNet \cite{movinets} produced by NAS, Video Mobile-Former achieves better results with fewer FLOPs on Kinetics-400 and Kinetics-600. Video Mobile-Former-5G also outperforms X3D-XS with less than $1/4$ total FLOPs. Since almost all video transformers developed in the previous literature have huge computational costs (\eg, more than 300 GFLOPs), we reimplement previous state-of-the-art video transformers, MViT and Video Swin Transformer, at a low FLOP regime. As shown in Table~\ref{tab:k400}, the variants of MViT and Video Swin suffer significant performance degradation when the computational budget is less than 50 GFLOPs, and Video Mobile-Former beats them in terms of accuracy even with $75\times$ smaller FLOPs. This suggests that Video Mobile-Former is much more efficient and effective than previous video transformers when the computational cost is limited.

\vspace{0.05in}
\noindent \textbf{Comparisons with efficient 3D-CNNs.} 
By using large temporal stride ($s_t = 8$), we reduce the computation of Video Mobile-Former under 1 GFLOPs. The previous work~\cite{kopuklu2019resource} also construct video models with extremely low FLOPs by extending various resource efficient 2D CNNs to 3D CNNs, and we compare Video Mobile-Former-560M with them on Kinetics-600, Diving48, UCF-101 and HMDB-51. The results in Table~\ref{tab:downstream} indicate that Video Mobile-Former achieves significantly higher accuracy on all datasets using similar computation measured by FLOPs.

\begin{table}[t]
  \centering
\setlength{\tabcolsep}{0pt} 
  \begin{tabular*}{\linewidth}{@{\extracolsep{\fill}}lccc@{}}
    \toprule
    Token Num & Top-1 & Top-5 & GFLOPs\\
    \midrule
    2 & 70.3 & 88.9 & 2.51  \\
    4 & 70.8 & 89.1 & 2.56  \\
    6 & 71.1 & 89.3 & 2.61  \\
    8 & 70.6 & 89.2 & 2.65 \\
    \midrule
    $T \times 2$ & 68.8 & 87.4 & 2.55  \\
    $T \times 4$ & 69.8 & 88.5 & 2.65  \\
    $T \times 6$ & 71.4 & 89.3 & 2.75  \\
    $T \times 8$ & 70.4 & 88.2 & 2.86 \\
    \bottomrule
  \end{tabular*}
    \vspace{-0.5em}
    \caption{Ablation study on the number of global tokens. Video Mobile-Former-2G is used and $T=16$ here.}
  \label{tab:token_num}
  \vspace{-1em}
\end{table}

\begin{table}[t]
  \centering
\setlength{\tabcolsep}{0pt} 
  \begin{tabular*}{\linewidth}{@{\extracolsep{\fill}}lccc@{}}
    \toprule
    Temporal Stride & Top-1 & Top-5 & GFLOPs \\
    \midrule
    1 & 60.3 & 82.6 & 1.43 \\
    2 & 63.7 & 84.7 & 1.51 \\
    4 & 67.4 & 87.4 & 1.43  \\
    8 & 65.0 & 85.1 & 1.44 \\
    \bottomrule
  \end{tabular*}
    \vspace{-0.5em}
  \caption{Ablation study on the temporal stride.}
  \label{tab:stride}
  \vspace{-1.5em}
\end{table}

\begin{table}[t]
  \centering
\setlength{\tabcolsep}{0pt} 
  \begin{tabular*}{\linewidth}{@{\extracolsep{\fill}}lcccc@{}}
    \toprule
    Temporal Downsampling & Top-1 & Top-5 & GFLOPs \\
    \midrule
    Each stage & 61.9 & 83.0 & 1.14 \\
    Stem & \textbf{62.6} & 83.7 & 0.56 \\
    \bottomrule
  \end{tabular*}
    \vspace{-0.5em}
  \caption{Ablation study on the position of temporal downsampling in Video Mobile-Former.}
  \label{tab:downsampling}
  \vspace{-1em}
\end{table}

\begin{table}[t]
  \centering
\setlength{\tabcolsep}{0pt} 
  \begin{tabular*}{\linewidth}{@{\extracolsep{\fill}}lcccc@{}}
    \toprule
    Model & Depth & Top-1 & Top-5 & GFLOPs \\
    \midrule
    Mobile3D & 8 & 60.9 & 83.7 & 1.26 \\
    Video-Mobile-Former & 8 & 67.4 (+6.5) & 87.4 & 1.43 \\
    \midrule
    Mobile3D & 16 & 65.7 & 87.1 & 1.90 \\
    Video-Mobile-Former & 16 & 70.3 (+4.6) & 89.3 & 2.14 \\
    \midrule
    Mobile3D & 24 & 68.8 & 88.7 & 2.53 \\
    Video-Mobile-Former & 24 & 71.8 (+3.0) & 89.9 & 2.85 \\
    \midrule
    Mobile3D & 32 & 70.0 & 89.5 & 3.17 \\
    Video-Mobile-Former & 32 & 71.9 (+1.9) & 89.9 & 3.56 \\
    \bottomrule
  \end{tabular*}
    \vspace{-0.5em}
  \caption{Ablation study on the depth of networks.}
  \label{tab:depth}
  \vspace{-1.5em}
\end{table}

\begin{table}[t]
  \centering
\setlength{\tabcolsep}{0pt} 
  \begin{tabular*}{\linewidth}{@{\extracolsep{\fill}}lccc@{}}
    \toprule
    Conv Type & Top-1 & Top-5 & GFLOPs \\
    \midrule
    2D & 69.7 & 88.5 & 2.46 \\
    (2+1)D & 68.3 & 86.3 & 2.48 \\
    3D & \textbf{71.1} & \textbf{89.3} & 2.61 \\
    \bottomrule
  \end{tabular*}
    \vspace{-0.5em}
   \caption{Ablation study on the type of depthwise convolution in Mobile block. Video Mobile-Former-2G is used here.}
  \label{tab:conv_type}
  \vspace{-1em}
\end{table}

\begin{table}[t]
  \centering
\setlength{\tabcolsep}{0pt} 
  \begin{tabular*}{\linewidth}{@{\extracolsep{\fill}}lcccc@{}}
    \toprule
    Activation & Top-1 & Top-5 & GFLOPs \\
    \midrule
    ReLU & 68.6 & 87.7 & 2.58 \\
    DY-ReLU & 69.3 & 87.9 & 2.58 \\
    Frame-level DY-ReLU & \textbf{71.1} & \textbf{89.3} & 2.61 \\
    \bottomrule
  \end{tabular*}
    \vspace{-0.5em}
  \caption{Ablation study on the activation function in Mobile block. Video Mobile-Former-2G is used here.}
  \label{tab:activation}
  \vspace{-1em}
\end{table}

\subsection{Ablation Studies}

\noindent \textbf{Ablation on the number of global tokens.}
In Video Mobile-Former, we use $M$ global tokens for a whole video instead of $M$ tokens for each frame respectively. To demonstrate that a few global tokens are enough for global spatial-temporal modeling, in Table~\ref{tab:token_num} we evaluate Video Mobile-Former using $M$ global tokens or $T \times M$ global tokens on Kinetics-400, where $M$ ranges from 2 to 8 and $T=64/4=16$. For both two methods of global spatial-temporal modeling, the biggest performance improvement is achieved by setting $M=6$. Moreover, the model using 6 global tokens achieves comparable top-1 accuracy with the model using $T \times 6$ tokens, but consumes less computational cost. This ablation shows that extending the global tokens on the temporal axis yields little performance gain and very few global tokens to model global information of the entire video is both sufficient and efficient.

\vspace{0.05in}
\noindent \textbf{Ablation on the temporal stride.}
To design a variant of Video Mobile-Former with lower FLOPs, one way is to use larger temporal stride while another way is to reduce the number of layers or channels. Table~\ref{tab:stride} shows the results for different temporal strides in the stem of models. When the temporal stride increases, we increase the width or the depth of Video Mobile-Former to keep the FLOPs almost the same. We find that when the temporal stride is set to 4, the best balance of the accuracy and the computational cost is achieved. 

\vspace{0.05in}
\noindent \textbf{Ablation on the position of temporal downsampling.}
Early work on video recognition~\cite{i3d,c3d} applies progressive temporal downsampling in different network stages. In contrast, we utilize large temporal downsampling directly in the stem of models, which substantially reduces the temporal length of feature maps from the early stage. To ablate the position of temporal downsampling, we design a variant of Video Mobile-Former which applies temporal downsampling at each stage in a progressive way. The results in Table~\ref{tab:downsampling} demonstrate that our default design achieves better performance with less computation, while other details of model architectures are the same.

\vspace{0.05in}
\noindent \textbf{Ablation on the depth of networks.}
The results in Table~\ref{tab:compare_cnn} demonstrate that global spatial-temporal modeling in Video Mobile-Former is an efficient and effective way to strengthen 3D-CNN models. To study the relationship between the depth of networks and the improvement of strengthening 3D CNNs with Video Mobile-Former architecture, Table~\ref{tab:depth} shows the performance gain of Video Mobile-Former with different depths compared with Mobile3D baselines. When the network becomes shallower, the gap between Video Mobile-Former and the Mobile3D baseline enlarges. This highlights that our Video Mobile-Former is more satisfactory for building extremely lightweight video models.

\vspace{0.05in}
\noindent \textbf{Ablation on the type of depthwise convolution.}
To determine whether 3D convolution is required when adapting Mobile-former to the video domain, we compare different convolution types for the depthwise convolution in Mobile blocks and compare the results in Table~\ref{tab:conv_type}. It is worth noting that we still use (2+1)D convolutions in the stem of these model variants to extract spatiotemporal features. It shows that the best performance is achieved by using 3D convolution in Video Mobile-Former. Since the computational cost of the depthwise convolution accounts for a small proportion of the overall computational cost of Video Mobile-Former, compared with 2D convolution, 3D convolution can achieve a significant performance improvement while requiring slightly more computational resources.

\begin{table}[!ht]
  \centering
\setlength{\tabcolsep}{0pt} 
  \begin{tabular*}{\linewidth}{@{\extracolsep{\fill}}lccc@{}}
    \toprule
    Model & On GPU & On CPU & K400 \\
      & (ms) & (ms) & Top-1 \\
    \midrule
    Video-Mobile-Former-560M & 26.1 & 192.1 & 62.6 \\
    Video-Mobile-Former-1G & 29.0 & 603.5 & 67.4 \\
    Video-Mobile-Former-2G & 32.9 & 937.2 & 71.1 \\
    Video-Mobile-Former-5G & 57.9 & 1855 & 72.9 \\
    \midrule
    X3D-S & 26.9 & 5117 & 70.8 \\
    MViT-43G & 73.1 & 6820 & 62.0  \\
    MViT-80G & 82.6 & 8492 & 69.8  \\
    VideoSwin-212G & 60.8 & 6948 & 71.7 \\
    \bottomrule
  \end{tabular*}
    \vspace{-0.5em}
  \caption{Runtime on an Nvidia V100 GPU or an Intel Xeon Platinum 8168 CPU @ 2.70GHz on Kinetics-400. Latency is given for a video clip of 64 frames.}
  \label{tab:runtime}
  \vspace{-1.5em}
\end{table}

\vspace{0.05in}
\noindent \textbf{Ablation on the activation function.}
Table~\ref{tab:activation} shows the results on K400 for different activation function in the Mobile block. Using the frame-level CNN features and the first global token to generate parameters for dynamic ReLU gains 2.5\% top-1 accuracy over the baseline using the simple ReLU function and 1.8\% top-1 accuracy over the original design in Mobile-Former. This indicates that frame-level DY-ReLU is a better adaption of dynamic ReLU from the image domain to the video domain.

\vspace{0.05in}
\noindent \textbf{Runtime comparison during inference.}
In Table~\ref{tab:runtime}, we first measure the inference speed of Video Mobile-Formers on an Nvidia V100 GPU, and compare our models with other efficient video models. Note that we evaluate latency on 64 frames for all models. When running on a single GPU, the latency of Video Mobile-Former is comparable to that of the X3D baseline, and our models can run faster than video transformers with a similar top-1 accuracy. We also test the real-time latency of different models on an Intel Xeon Platinum 8168 CPU @ 2.70GHz. We find that the CPU latency of our models is much less than both 3D-CNN baselines and video transformers, which is more friendly for mobile devices. 

\section{Conclusion}

Video Mobile-Former is an efficient and effective extension of Mobile-Former based on efficient global spatial-temporal modeling. It enables global interaction with very few global tokens. To achieve better balance of accuracy and computational cost for video recognition, in consideration of the frame redundancy of videos, we leverage only 6 global tokens to model the global relationship of a whole video and use large temporal stride in the stem of our model. Our proposed Video Mobile-Former achieves significantly better performance compared with 3D CNN baselines. On video recognition tasks, it outperforms both state-of-the-art efficient 3D CNNs and video transformers at the low FLOP regime. 

\bigskip
\bibliography{aaai23}


\end{document}